\documentclass[10pt,twocolumn,letterpaper]{article}

\usepackage{wacv}              % To produce the CAMERA-READY version
\usepackage[accsupp]{axessibility} % Improves PDF readability for those with disabilities
\usepackage{graphicx}
\usepackage{amsmath}
\usepackage{amssymb}
\usepackage{booktabs}
\usepackage{multirow}
\usepackage{makecell}
\usepackage{amssymb} % For the checkmark symbol
\usepackage{array}
\usepackage{booktabs}
\usepackage{tabularx}

\usepackage[pagebackref,breaklinks,colorlinks]{hyperref}

\usepackage[capitalize]{cleveref}
\crefname{section}{Sec.}{Secs.}
\Crefname{section}{Section}{Sections}
\Crefname{table}{Table}{Tables}
\crefname{table}{Tab.}{Tabs.}

\def\wacvPaperID{00010} % *** Enter the WACV Paper ID here
\def\confName{WACV}
\def\confYear{2025}

\begin{document}

%%%%%%%%% TITLE - PLEASE UPDATE
% \title{Feature Augmentation-Driven Contrastive Representations for Few-Shot Class Incremental Learning}

\title{\vspace{-2.5em} Strategic Base Representation Learning via Feature Augmentations for Few-Shot Class Incremental Learning \vspace{-1em}}

% \author{Parinita Nema\\ 
% IISER Bhopal \\
% India\\
% {\tt\small parinita22@iiserb.ac.in}
% % For a paper whose authors are all at the same institution,
% % omit the following lines up until the closing ``}''.
% % Additional authors and addresses can be added with ``\and'',
% % just like the second author.
% % To save space, use either the email address or home page, not both
% \and
% Vinod K Kurmi\\
% IISER Bhopal\\
% India\\
% {\tt\small vinodkk@iiserb.ac.in}
% }
% \maketitle

\author{Parinita Nema, Vinod K Kurmi\\ 
Indian Institute of Science Education and Research Bhopal, India \\
{\tt\small \{parinita22, vinodkk\}@iiserb.ac.in}}
% For a paper whose authors are all at the same institution,
% omit the following lines up until the closing ``}''.
% Additional authors and addresses can be added with ``\and'',
% just like the second author.
% To save space, use either the email address or home page, not both

\maketitle

%%%%%%%%% ABSTRACT

\begin{abstract}
Few-shot class incremental learning implies the model to learn new classes while retaining knowledge of previously learned classes with a small number of training instances. Existing frameworks typically freeze the parameters of the previously learned classes during the incorporation of new classes. However, this approach often results in suboptimal class separation of previously learned classes,  leading to overlap between old and new classes. Consequently, the performance of old classes degrades on new classes. To address these challenges, we propose a novel feature augmentation driven contrastive learning framework designed to enhance the separation of previously learned classes to accommodate new classes. Our approach involves augmenting feature vectors and assigning proxy labels to these vectors. This strategy expands the feature space, ensuring seamless integration of new classes within the expanded space. Additionally, we employ a self-supervised contrastive loss to improve the separation between previous classes. We validate our framework through experiments on three FSCIL benchmark datasets: CIFAR100, miniImageNet, and CUB200. The results demonstrate that our Feature Augmentation driven Contrastive Learning framework significantly outperforms other approaches, achieving state-of-the-art performance. Code is available at \href{https://visdomlab.github.io/FeatAugFSCIL/}{{https://visdomlab.github.io/FeatAugFSCIL/}}.
\end{abstract}

%%%%%%%%% BODY TEXT
\vspace{-1em}
\section{Introduction}

Deep learning has demonstrated remarkable performance across various computer vision tasks~\cite{miyato2018spectral, he2017mask}. This superior performance is achieved due to the availability of static and large volumes of annotated training data. However, in practical applications, data is often received in a non-static, continuous stream, and there is typically a limited amount of labeled training data available. For example, facial recognition systems often require incrementally identifying new faces with a small number of labeled samples while also retaining information about previously learned faces.
\begin{figure}[h!]
  
  % \vspace{-10pt} % Adjusts space above the image
  \includegraphics[height=0.28\textwidth]{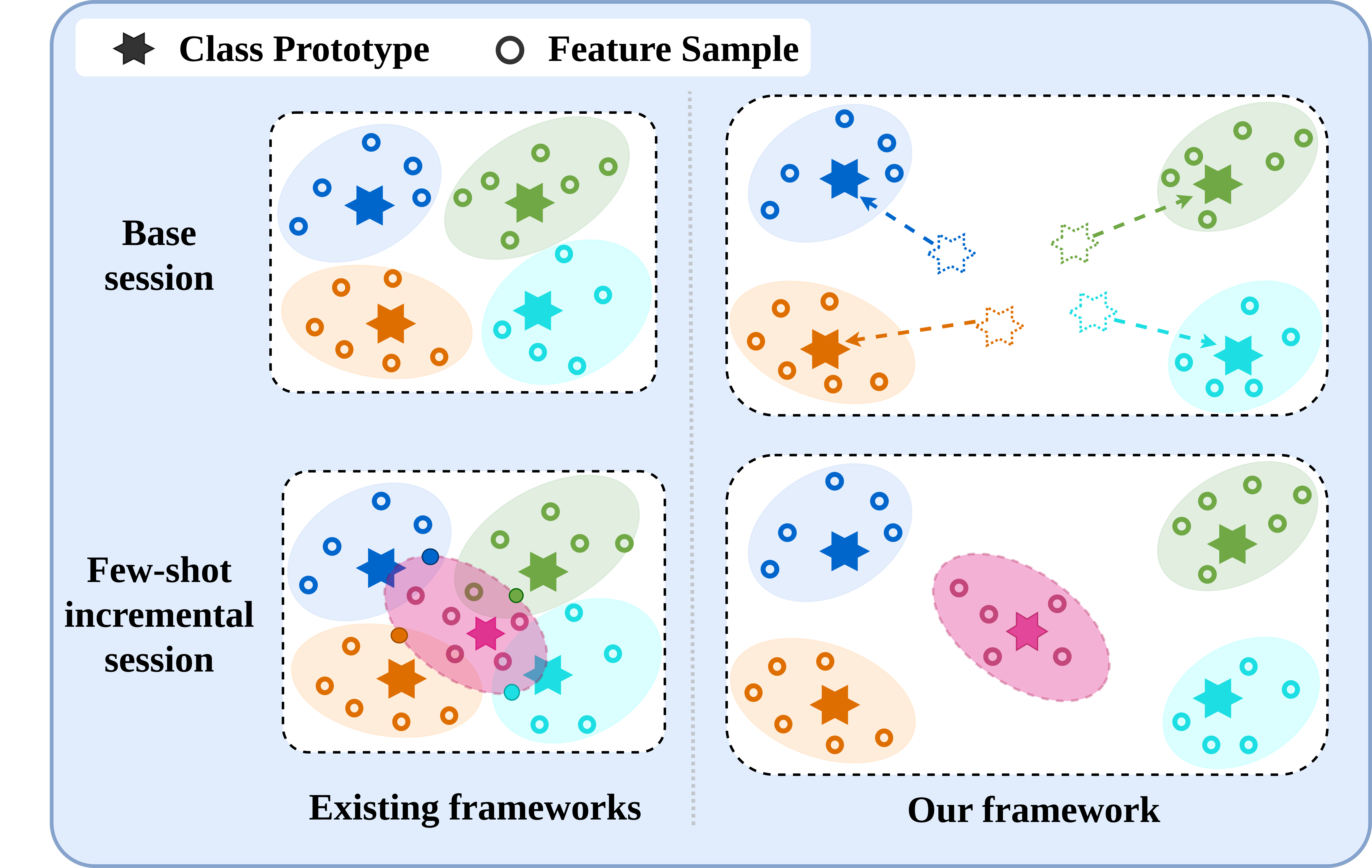}
  \centering
  \caption{The illustration of the motivation for our framework. Our FACL framework is designed to more effectively learn the base session features and facilitate future class adaptation compared to existing approaches.}
  \label{fig:1}
  \vspace{-10pt} % Adjusts space below the caption
\end{figure}

Class incremental learning (CIL)~\cite{hou2019learning, li2017learning, rebuffi2017icarl, yan2021dynamically,kurmi2021not} was introduced, enabling the system to learn new classes continuously while preserving previously acquired knowledge. Few-Shot Class-Incremental Learning (FSCIL) has gained significant interest~\cite{tao2020few, chen2020incremental, dong2021few, shi2021overcoming, zhou2022forward, singh_intespeech24}. FSCIL requires the model to classify new data while keeping the knowledge it has previously learned despite having only a few labeled samples for new classes. In FSCIL, there are sufficient samples available during the initial training session, but only a limited number of samples for the newly introduced classes during incremental learning phases.

The primary challenges of FSCIL are catastrophic forgetting and overfitting~\cite{chen2020incremental, zhao2021mgsvf, dong2021few}. Catastrophic forgetting occurs when the model, while learning new classes without access to data from previously learned classes, forgets what it has previously learned due to weight drift. Additionally, because the new classes have only a few labeled samples, the model is prone to overfitting on these new classes.

To address these challenges, recent methods~\cite{zhang2021few, zhao2023few, zhu2021self} involve training the model during the base session and then freezing the backbone to facilitate learning on the novel classes. Although a fixed backbone maintains performance on the base session, it does not provide optimal generalization for the incremental sessions. So, it is essential to consider the following question: \textit{How can the training of the base session model be improved to ensure it can effectively adapt to future classes?} In the base session, if all the representations are well-separated, incremental classes can be incorporated with ease, avoiding any overlap with the base classes as represented in Figure~\ref{fig:1}.

In this work, we propose a method called Feature Augmentation driven Contrastive Learning (FACL). Our method organizes the feature space of the base session in a manner that enhances the feature adaptability of the incremental session classes. To achieve this, we employ feature augmentation and the creation of proxy classes to effectively structure the feature space for incremental session adaptation. This augmented feature sets allow the classifier to maintain the decision boundaries for the classes it has learned so far. Moreover, proxy classes are created for the augmented features to ensure they allocate sufficient space for incremental classes.

% By utilizing feature augmentation, the model will be able to classify unknown classes in advance and reserve space for future classes.
Feature augmentation enhances the robustness of image features, facilitating better separation of the base classes. For feature augmentation, we use a mixture of two different augmented images to generate representations for proxy classes. By utilizing these diverse representations, the features of the base classes are more distinct from each other, effectively expanding the feature space. Subsequently, the model contrasts the representations of base classes by incorporating a self-supervised contrastive loss into our framework. This ensures the separation of base classes, and enhances accuracy by incorporating contextual information. Our augmented features provide more compact representations for proxy classes, increasing the distance between base classes and facilitating the adaptation of new classes. Our main contributions are summarized in three key points: 
\begin{itemize}
    \setlength{\itemsep}{0pt} % Adjust the item separation
    \setlength{\parskip}{0pt} % Adjust the paragraph separation
    \item We introduce a novel feature augmentation driven contrastive learning framework, which involves feature augmentation and proxy class generation to organize the feature space in advance for future class adaptation.
    \item We employ a self-supervised contrastive loss to enhance the separation between the base classes.
\item  We provide a thorough evaluation of the proposed framework by considering detailed comparison on standard benchmark datasets  such as CIFAR100, \textit{mini}ImageNet, and CUB200 and achieve average \textbf{4.77\%}, \textbf{0.6\%} and \textbf{1.59\% }improvement against state-of-the-art methods.
    % \item Experiments are conducted on FSCIL benchmark datasets: CIFAR100, \textit{mini}ImageNet, and CUB200. The results demonstrate that the FACL framework significantly outperforms other approaches, achieving state-of-the-art (SOTA) performance.
\end{itemize}

\section{Related Work}
\subsection{Few-Shot Learning}
Few-shot learning aims to train a model with a very limited number of labeled training samples. Methods of few-shot learning, trained with a limited number of samples, encounter the problems of overfitting and data imbalance. Current methods fall into two main categories: metric-based methods~\cite{oreshkin2018tadam, snell2017prototypical, vinyals2016matching} and optimization-based methods~\cite{finn2017model, nichol2018first, qi2018low}. In our work, we primarily focus on metric-based methods for few-shot learning. In metric-based methods, an initial training is conducted on a large dataset to learn the similarity metric. Subsequently, the similarity score is calculated between the support and query sets. Optimization-based methods employ meta-learning architectures to determine optimal parameters, enabling the model to Acquire generalize representations.
\subsection{Class Incremental Learning}
In CIL, data is continuously streamed, requiring the model to classify new categories while retaining knowledge of previously learned categories. CIL algorithms can be predominantly classified into three categories. The first category includes replay-based methods~\cite{rebuffi2017icarl, belouadah2019il2m, hou2019learning, zhu2021prototype}, which store a portion of samples from earlier training sessions to be used in later incremental sessions, thus reducing the forgetting of previous categories. The second category comprises regularization-based methods~\cite{li2017learning, liu2018rotate}, which adjust model parameters to minimize forgetting by penalizing changes to important parameters. The third category involves architecture-based methods~\cite{yan2021dynamically, zhu2021class}, where the model's ability to classify new categories is enhanced by dynamically expanding the network architecture. ORTHOG-SUBSPACE~\cite{chaudhry2020continual} offers a method where tasks are learned in orthogonal vector subspaces and projected tasks into lower-dimensional spaces using task-specific projection matrices. BatchEnsemble~\cite{wen2020batchensemble} generates the weights through the Hadamard product; this gives efficient parallel processing across and within devices. TKIL~\cite{xiang2023tkil}, on the other hand, introduces the GTK (gradient tangent loss)  that reduces the difference between the old and new gradient of the task. Our approach aligns with this third category.
\subsection{Few-Shot Class-Incremental Learning}
FSCIL addresses the challenge of training CIL tasks with a limited number of samples. FSCIL initially introduced by TOPIC~\cite{tao2020few}, To prevent forgetting and preserve feature topology, TOPIC employs a neural gas network. Following this, To improve model performance, CEC~\cite{tao2020few} used pseudo incremental sessions to train an attention-based module. FACT~\cite{zhou2022forward} introduced virtual prototypes to allocate space for future classes. NC-FSCIL~\cite{yang2023neural} adopted the concept of neural collapse, aligning the final layer features and classifier prototypes into an optimal geometric configuration. ALICE~\cite{peng2022few} utilizes the angular penalty loss to learn sparse feature representations for future classes, while employing mixup for data augmentation and generating synthetic auxiliary class data. Meanwhile, MICS~\cite{kim2024mics} introduces the midpoint mixup technique, leveraging soft labels to regulate the probability of mixup images and providing more control over the data synthesis process. By using "Incrementally Learnt Angular Representations," ILAR~\cite{yoon2023incrementally} raises the margin of the features of the class and introduces a feature generator for incremental session training. In contrast, our method generates pseudo classes using a combination of feature vectors and employs contrastive loss to improve the generalization of novel classes. 

\vspace{-0.5em}
\section{Methodology}
\subsection{Preliminary}

FSCIL training is conducted in two modes: 1) Base session training and 2) Incremental session training. During the base session, a sufficient amount of training data is available. In the incremental sessions, data is introduced incrementally in a few-shot manner. The sequence of training data is represented as $\{\mathcal{D}_{\text{tr}}^{(0)}, \mathcal{D}_{\text{tr}}^{(1)}, \ldots, \mathcal{D}_{\text{tr}}^{(I)}\}$, for any given session $s$, $\mathcal{D}_{\text{tr}}^{(s)} = \{ (x_i, {y}_i) \}_{i=1}^{|\mathcal{D}_{\text{tr}}^{(s)}|}$, here $\mathcal{D}_{\text{tr}}^{(0)}$ denotes the training data of the base session and $I$ represents the total number of incremental sessions and $x_i$ is the \( i^\text{th} \) image and ${y}_i$ is the corresponding label of image $x_i$. Incremental sessions are denoted as $\{\mathcal{D}_{\text{tr}}^{(s)} \, | \, s > 0 \}$ where $|\mathcal{D}_{\text{tr}}^{(s)}| = mn $, here $m$ is the number of classes and $n$ is the total number of samples in each class. For any specific $s^{\text{th}}$ session, the label space is denoted as $\mathcal{Y}^{(s)}$. In the FSCIL setup, there should be no overlap of the classes between sessions, i.e., $\mathcal{Y}^{(s)} \cap \mathcal{Y}^{(s')} = \emptyset$ for all $s' \neq s$. The model trained on the training data $\mathcal{D}_{\text{tr}}^{(s)}$ is evaluated on the test data $\mathcal{D}_{\text{ts}}^{(s)}$. Testing is performed on all encountered classes up to the $s^{\text{th}}$ session, i.e., $\mathcal{Y}^{(0)} \cup \mathcal{Y}^{(1)} \cup \ldots \cup \mathcal{Y}^{(s)}$.

\subsection{Expanding Base Class Feature Space Facilitates Future Class Adaptation}

\begin{figure*}[h]
  \centering
  % \resizebox{\textwidth}{!}{\includegraphics{Copy of audio classification (1).png}}
  \includegraphics[height=0.45\textwidth]{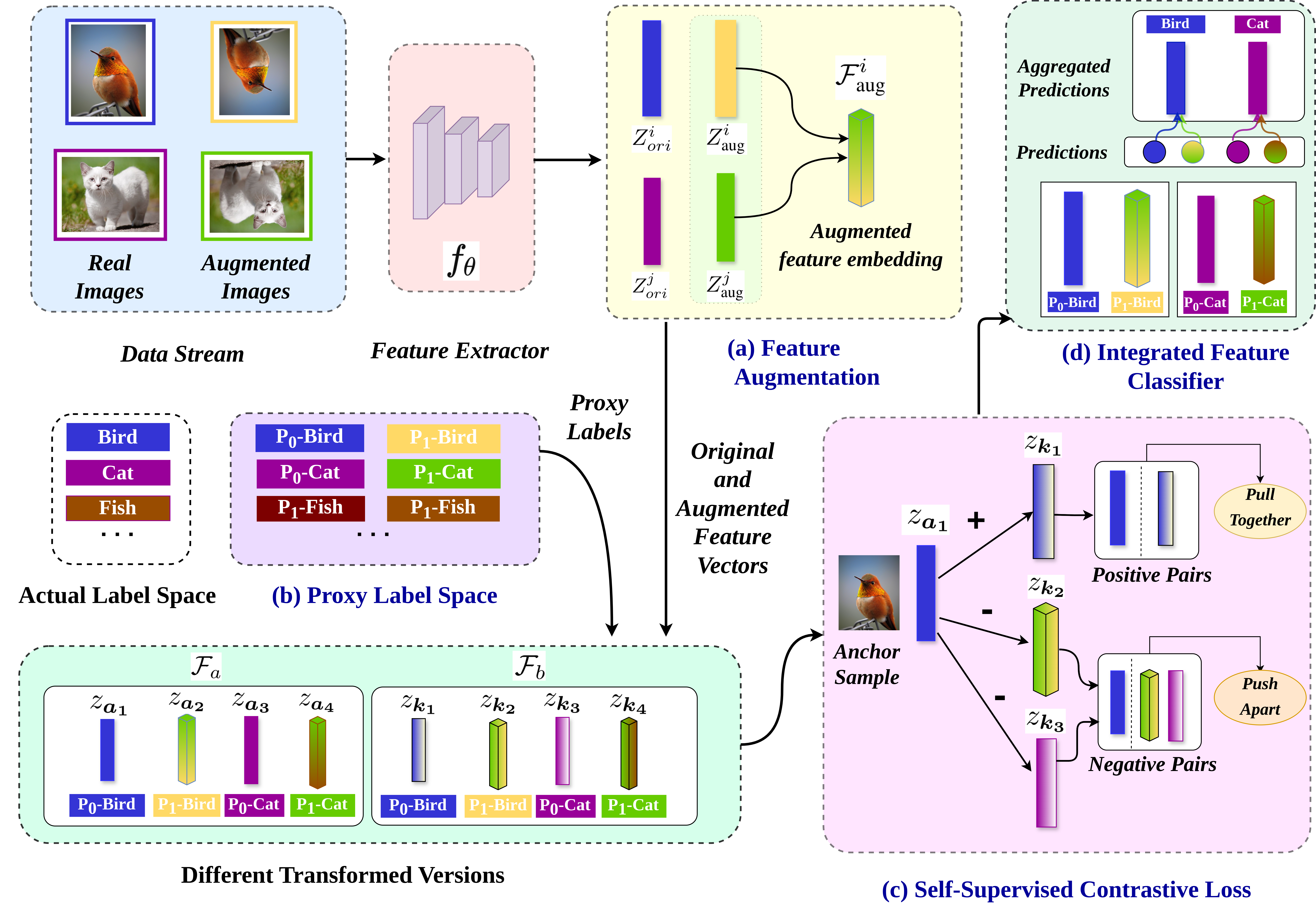} 
  \vspace{-1em}
  \caption{The illustration of the proposed FACL framework. FACL consists of three main components: (a) feature augmentation, where the augmented feature vectors are used to perform the feature augmentation, (b) proxy label space, where proxy labels are created for the augmented feature vectors, and (c) self-supervised contrastive loss, where contrastive loss is applied to the different transformed versions $\mathcal{F}_a$ and $\mathcal{F}_b$, which are created using $\mathcal{F}_{\text{comb}}$ as defined in Eq.~\ref{eq:combined_features} and built using the original image feature vectors and the augmented feature vectors. (d) Integrated Feature Classifier, where at inference time, the original label logits and the proxy label logits are integrated to perform the evaluation.}
  % \vspace{-em}
  \label{fig:method}
  \vspace{-1.76em}  
\end{figure*}

In previous works~\cite{shi2021overcoming, wang2024few, zhou2022forward}, frozen pretrained feature extractor methods have demonstrated that pre-training during the base session is crucial for effectively accommodating novel classes in subsequent incremental sessions.  The performance of these prior methods, which utilize a frozen backbone, motivates us to develop a more robust base session to ensure that novel classes are easily accommodated without overlapping with base classes. We consider how our base session training should be conducted to incorporate future incremental class adaptation. 
We hypothesize that maximizing the separability of base class decision boundaries can enhance future adaptation to few-shot classes. If the base session feature space is expanded, there will be sufficient room available for future class adaptation. Consequently, the base class features are learned in a manner that prevents confusion with incremental class features. \\
During the base session training of FSCIL, a frozen feature extractor and classifier are typically used, with the model trained on sufficient data using the cross-entropy loss. The model has two main components: a feature extractor and a classifier. The feature extractor is represented as \( f_\theta \), where \(\theta\) denotes the parameters of the model. The input image \( x_i \) is mapped to a \(\mathbb{R}^D\) space, where \(D\) signifies the dimensions of the feature vector. For $\mathcal{Y}$ be the label space, and let $C$ be a classifier that assigns a label $y \in \mathcal{Y}$ to an input $x$. For each class $y \in \mathcal{Y}$, there is a classifier $c_y$. The classifier loss $\mathcal{L}$ for an input-output pair $(x, y)$ can be defined as follows:
\begin{equation}
\mathcal{L}_{\text{classifier}}(\varphi ; {x}, y) = \mathcal{L}_{\text{CE}}(\varphi({x}), y)
\end{equation}
where \(\mathcal{L}_{\text{CE}}\) is the cross-entropy loss, and \(\varphi({x})\) represents the combination of the feature extractor and the classifier. In each incremental session, the classifier \( C \) is extended to accommodate new classes. The classifier \( C \) is defined as:
$C = \{ c^0_1, c^0_2, \ldots, c^0_{|\mathcal{Y}^{(0)}|} \} \cup \ldots \cup \{ c^s_1, c^s_2, \ldots, c^s_{|\mathcal{Y}^{(s)}|} \}$,  where \(|\mathcal{Y}^{(s)}|\) represents the number of classes in the $s^{\text{th}}$ incremental session. This extension allows the classifier to incorporate the new classes introduced during each incremental session. 
Recent works~\cite{zhao2023few, wang2024few} primarily focus on the conventional cross-entropy loss. However, this approach alone is insufficient to achieve optimal performance in future incremental learning sessions. To enhance the robustness of the base session model, we have proposed a straightforward yet effective approach known as Feature Augmentation driven Contrastive Learning (FACL). Our approach comprises two primary components: \textit{Feature augmentation} and \textit{Self-supervised contrastive learning (SSCL).} Our comprehensive pipeline is illustrated in Figure \ref{fig:method}, showcasing the integrated processes and components involved.

\subsection{Feature Augmentation driven Contrastive Learning}
% The analysis that we have conducted clearly indicates that existing methods lead to poor generalization of novel classes. This results in significant overlap between base and novel classes, causing confusion between them.
Existing methods tend to result in poor generalization for novel classes, leading to significant overlap with base classes and causing confusion. The confusion arises because the features of the base session lack adequate diversity, leading to a decision boundary that is overly precise and closely fitted to the training data, and when training on incremental sessions, the decision boundaries from previous tasks change. So in the base session, the model should be trained in a manner that incorporates all necessary preparations and adjustments in advance, based on what might occur in the future. The challenge of class overlap and facilitation of novel class adaptations is addressed by our method, FACL. Initially, we employ a feature augmentation approach to enhance the ability of the model for future class adaptation. Next, the model is trained with SSCL to achieve better separation of the base session classes. \\

% In our training pipeline, the base session training consists of two main stages. In the first stage, the training batch is augmented, resulting in both original and augmented datasets. Let $\mathcal{B}_{\text{tr}}$ represent the original training batch and $\mathcal{B}_{\text{aug}}$ the augmented batch. Features of both the original and augmented datasets are extracted. The feature embeddings for the original and augmeneted batches are $f_\theta(\mathcal{B}_{\text{tr}})$, and $f_\theta(\mathcal{B}_{\text{aug}})$. Then, we use these augmented dataset features to perform feature augmentations and form a combined feature set $\mathcal{F}_{\text{comb}}$. In the second stage, the base session training is conducted using an unsupervised contrastive loss on the combined feature set. Finally, we apply a cross-entropy loss for classification, combining both losses to enhance the model's performance.

\vspace{-2em}
\subsubsection{Feature Augmentation}

The proposed approach, FACL aims to enhance the base session training by incorporating feature augmentation. This process is designed to facilitate better adaptation to future classes.

The training dataset for the base session is denoted as \( \mathcal{D}_{\text{tr}}^{(0)} \). For a given mini-batch \(\{ x_1, x_2, \ldots, x_B \}\) where \( B \) represents the mini-batch size, a set of transformations \( \mathcal{M} \) is applied to obtain augmented versions of each image \( x_i \). The augmented training dataset is denoted as \( \mathcal{D}_{\text{tr}}^{\text{Aug}(0)} \), comprising the augmented mini-batch \(\{x'_1, x'_2, \ldots, x'_{B'}\}\). Here, \( B' \) is the size of the augmented mini-batch, determined by the number of transformations applied, specifically \( B' = B \times (\mathcal{M} + 1) \).

In previous methods such as CEC~\cite{zhang2021few} utilize attention mechanisms and train encoders using feature space-based methods. In this context, we introduce an effective approach called feature augmentation to expand the feature space for incremental session classes. The feature extractor \( f_{\theta} \) extracts features for both the original training dataset \( \mathcal{D}_{\text{tr}}^{(0)} \) and the augmented training dataset \( \mathcal{D}_{\text{tr}}^{\text{Aug}(0)} \). The feature vectors for the original and augmented datasets are denoted as \( Z_{\text{ori}} \) and \( Z_{\text{aug}} \), respectively, where:
\vspace{-0.5em}
\begin{equation}
\vspace{-0.5em}
% \begin{aligned}
Z_{\text{ori}} = f_{\theta}(\mathcal{D}_{\text{tr}}^{(0)}); \hspace{2em}
Z_{\text{aug}} = f_{\theta}(\mathcal{D}_{\text{tr}}^{\text{Aug}(0)})
% \end{aligned}
\end{equation}

Our objective is to develop high-level semantic features to enhance the training of future samples. The core concept of the FACL approach is to generate a mixture of features that differ from the original ones. This mixture of features serves as a proxy for new incremental classes, thereby improving the base session training for future class adaptation. 

To construct the feature augmentation, we employ a method where we systematically blend two feature vectors from augmented images, denoted as \( Z^i_{\text{aug}} \) and \( Z^j_{\text{aug}} \), to generate a new feature vector \( \mathcal{F}^{i}_{\text{aug}} \). This newly generated feature vector serves as a proxy for a novel class. Here, \( Z^i_{\text{aug}} \) represents the \( i\textsuperscript{th}
 \) feature vector and \( Z^j_{\text{aug}} \) represents the \( j\textsuperscript{th}\) feature vector of the augmented images, specifically from the dataset \( \mathcal{D}^{\text{Aug}(0)}_{\text{tr}} \). The process involves three distinct steps: 1) Pairs of image feature vectors from \( Z_{\text{Aug}} \) are created. A random feature vector from another image in the batch is selected to form these pairs for a specific image feature vector. 2) To create an augmented feature vector, \( Z_{\text{aug}}(x_i) \) is combined with another random feature from \( Z_{\text{aug}}(x_j) \). The combination is performed using a \( \delta \) coefficient. 3) Finally, these new augmented features are incorporated into the training of the base session. The augmented feature vector is defined as follows:
\begin{equation}
\mathcal{F}^i_{\text{aug}} = \delta \cdot Z_{\text{aug}}^i + (1 - \delta) \cdot Z_{\text{aug}}^j
\label{eq:augmented_features}
\end{equation}

% The dimensions of the augmented feature vector \( F_{\text{aug}} \) are the same as those of the original feature vector \( Z_{\text{aug}} \), both being \( B \times \mathbb{R}^D \).

Formally, for each \( i \in \{1, 2, \ldots, B'\} \), a random index \( j \in \{1, 2, \ldots, B'\} \) is selected such that \( j \neq i \). To formalize the incorporation of the augmented features into the base session training with the updated notation, we define the combined feature vector \( \mathcal{F}^i_{\text{comb}} \) as follows:   
 \vspace{-0.6em}
\begin{equation}
\vspace{-0.7em}
\mathcal{F}^i_{\text{comb}} = \text{\textit{concat}}(Z^i_{\text{ori}}, \mathcal{F}^i_{\text{aug}})
% \vspace{-0.3em}
\label{eq:combined_features}
\end{equation}

Here, \(\text{\textit{concat}}(\cdot)\) denotes a concatenation function that interleaves each original feature vector from \( Z_{\text{ori}} \) with each corresponding augmented feature vector from \( \mathcal{F}_{\text{aug}} \). This interleaving is maintained such that the combined feature vectors are represented in the form $ \{ (q, q') \} $. The indices \( q \) and \( q' \) represent the positions of the feature vectors in their respective sets, where \( q \in \{0, 1, 2, \ldots \} \) for the original feature vectors and \( q' \in \{0, 1, 2, \ldots \} \) for the augmented feature vectors.
\vspace{-1em}
\subsubsection{Proxy Class Generation}
After feature augmentation, we further generate proxy classes for these augmented feature vectors. In FSCIL, access to future incremental classes is not available. However, we can create proxy classes during the base session. These proxy classes act as placeholders and allow us to expand our label space, facilitating the incorporation of future classes in subsequent incremental sessions. Consequently, for each new augmented feature vector, a new proxy target is generated. To generate the proxy classes, we perform the modification of our original label space \(\mathcal{Y}\). To transform the labels, the following formula is employed:
\vspace{-1em}
\begin{equation}
y_p = y \times \mathcal{P} + p
\end{equation}

\vspace{-0.70em}
% Here, we have taken \( M = 1 \), which means that for a particular image, we only take one augmentation. 
% Here, we have set \( M = 1 \) as described in Section 3.1.1, which means that for a particular image, we only apply one augmentation.

The label needs to be determined for both the original and augmented features. In this context, \( \mathcal{P} \) is the number of features vectors for which labels should be decided. Therefore, \( \mathcal{P} = 2 \), so for each particular label:
% \vspace{-1em}
\begin{equation}
y = \{y_p\}_{p=0}^\mathcal{M}
\vspace{-0.5em}
\end{equation}

Where \( \mathcal{M} \) is the number of transformations and \( y_p \) is the \( p^\text{th} \) proxy target for the augmented features. Thus, in this approach, proxy targets are generated for each set of augmented features. By generating these proxy classes, we expand the representation space by a factor of \( \mathcal{P} \). This allows us to create more fine-grained proxy classes for the newly augmented feature vectors. These proxy targets are useful for maximizing the margins between classes and preserving space for future classes, thereby preventing confusion between novel and base classes.
\vspace{-1.2em}
\subsubsection{ Self-Supervised Contrastive Learning}
Base session training proceeds following the generation of augmented features and proxy classes. SSCL~\cite{khosla2020supervised} is incorporated to enhance the separation. In this approach, the contrastive learning model is trained to learn representations such that similar samples are positioned closer together in the feature space, while dissimilar samples are positioned further apart. In our approach, we incorporate newly augmented feature vectors, $\mathcal{F}_{\text{comb}}$. This incorporation introduces additional space for incremental classes, facilitating the separation between classes. 

We utilize the MoCo~\cite{chen2020improved} architecture for implementing contrastive learning in our approach. Specifically, generate different versions of augmented feature vectors, $\mathcal{F}_{\text{comb}}$. We construct $\mathcal{F}_a, \mathcal{F}_b   = \mathcal{A}(\mathcal{F}_{\text{comb}})$, where $\mathcal{F}_a, \mathcal{F}_b$ are the two different augmented versions.  $\mathcal{A}(\cdot)$ represents a random image augmentation set. Additionally, utilize a feature and label queue~\cite{he2020momentum}, which stores the most recent feature embeddings and their labels. The per-sample contrastive loss is defined as:

\vspace{-1.5em}
\begin{equation}
\begin{aligned}
& \mathcal{L}_{sscont} (\theta; \mathcal{F}_a, \mathcal{F}_b, \tau, \mathcal{N}, \mathcal{M}) = \\
& -\frac{1}{\mathcal{P}} \sum_{p=1}^{\mathcal{P}} \frac{1}{\left|S(j)\right|} \sum_{\boldsymbol{k}^+ \in S(j)} \log \frac{\exp \left( \boldsymbol{z}_a \cdot \boldsymbol{z}_{\boldsymbol{k}^+} / \tau \right)}{\sum_{\boldsymbol{k}' \in \mathcal{N}(j)} \exp \left( \boldsymbol{z}_a \cdot \boldsymbol{z}_{\boldsymbol{k}'} / \tau \right)}
\end{aligned}
\end{equation}

Here, consider $\boldsymbol{z}_a$ as the anchor sample from $\mathcal{F}_a$, and $j \in \{1, 2, \ldots, N\}$ to be the index of $N$ features from $\mathcal{F}_b$. $|S(j)|$ represents the number of positive pairs for the anchor $\boldsymbol{z}_a$. $\boldsymbol{z}_{\boldsymbol{k}^+}$ is the augmented sample embedding that originates from the same anchor sample embedding. $\mathcal{N}(j)$ includes all the embeddings from $\mathcal{F}_b$ and the embeddings from the queue. $\tau$ is the temperature parameter, and $(\cdot)$ denotes the dot product. During the base session training, we process this contrastive loss, while for the classification, a classifier is trained using the cross-entropy loss. Here, our augmented features and proxy targets are used to classify the model. Thus, the loss is defined as:
\vspace{-0.2em}
\begin{equation}
\mathcal{L}_{class}(\varphi ; \mathcal{F}_a, y, \mathcal{\mathcal{P}})=\frac{1}{\mathcal{P}} \sum_{i=1}^\mathcal{P} \mathcal{L}_{C E}\left(\varphi\left(\mathcal{F}_a\right), y_p\right)
\end{equation}

In the base session training, the model is trained using the joint loss, which is defined as:
% \vspace{-1em}
\begin{equation}
\mathcal{L} =  \mathcal{L}_{class} +  \mathcal{L}_{sscont}
\end{equation}

Where $\mathcal{L}_{class}$ and $\mathcal{L}_{sscont}$ represent the cross-entropy and self-supervised contrastive loss, respectively. 

In an incremental session training setup, where there are only a few samples per class, feature augmentation and proxy class generation are performed to effectively update the classifier's weights. Specifically, the weight of the classifier for a particular class \( c \) in the $s^{\text{th}}$ session is updated using the class prototype, The updated weight \( \boldsymbol{w}_c^s \) of the classifier for class \( c \) in the $s^{\text{th}}$ session is given by:
\vspace{-0.65em}
\begin{equation}
\boldsymbol{c}_y^s = \frac{1}{n_y^s} \sum_{i=1}^{n_y^s} f\left({x}_{y,i}\right)
\end{equation}

Here, \( n_y^s \) denote the total number of samples in class \( y \) during the $s^{\text{th}}$ session. The $i^{\text{th}}$ feature vector for class \( y \) is denoted by \( f\left({x}_{y,i}\right) \).

\vspace{-1em}
\subsubsection{Integrated Feature Evaluation}

For evaluation, we employ the nearest-mean (NCM) algorithm to assess the accuracy in each session across all classes encountered so far. In this process, compute the distance between the feature embedding \( f({x}) \) and all class prototypes. Specifically, the class \( c_{{x}} \) for a given sample \( {x} \) is determined by maximizing the cosine similarity between the feature embedding and the prototypes, as described by the equation below. Here, \( \operatorname{sim}(\cdot) \) denotes the cosine similarity between two vectors.
\vspace{-0.5em}
\begin{equation}
e_{{x}} = \arg \max _{y, s} \operatorname{sim}\left(f({x}), \boldsymbol{c}_y^s\right),
\vspace{-0.5em}
\end{equation}

where we calculate the cosine similarity between the features and the prototype, so \( \operatorname{sim}(\cdot) \) denotes the cosine similarity between two vectors. For evaluation in our approach, aggregate the original label predictions and proxy label predictions to compute the accuracy for the original labels.

% To mitigate this confusion, we expand the representation space of the base session. 
\vspace{-0.5em}
\section{Results and Experiments}
\subsection{Experimental Setup} 
\noindent\textbf{{Datasets:}} To evaluate the FSCIL performance of the proposed methods, we conduct experiments on three benchmark datasets: CIFAR100~\cite{krizhevsky2009learning}, \textit{mini}ImageNet~\cite{russakovsky2015imagenet}, and CUB200~\cite{wah2011caltech}. Statistical details for all datasets are provided in Table \ref{tab:dataset}.

% -----------------------------------------------------------------------------
\begin{table}[h]
\centering
\scalebox{0.86}{ % Adjust the scaling factor as needed
\begin{tabular}{lcccccc}
\hline
Dataset & $\mathrm{C}_b$ & $\mathrm{C}_{inc}$ & ${session}$ & ${ways}$ & ${shots}$ \\
\hline
CIFAR100 \cite{krizhevsky2009learning} & 60 & 40 & 8 & 5 & 5 \\
\textit{mini}ImageNet \cite{russakovsky2015imagenet} & 60 & 40 & 8 & 5 & 5 \\
CUB200 \cite{wah2011caltech} & 100 & 100 & 10 & 10 & 5 \\
\hline
\end{tabular}
}
\caption{Statistical summary of the datasets. In this context, $\mathrm{C}_b$ and $\mathrm{C}_{inc}$ represent the total number of classes in the base session and incremental sessions, respectively. ${session}$ denotes the total number of sessions, ${ways}$ indicates the number of classes, and ${shots}$ refers to the number of samples per class in each incremental session.}
\label{tab:dataset}
\vspace{-1em}
\end{table}
% ----------------------------------------------------------------------------------

\noindent\textbf{{Implementation details:}} In our experiments, We follow the CEC\cite{zhang2021few} protocol to conduct our experiments, utilize specific backbone models for training.  For the CUB200 and \textit{mini}ImageNet datasets, we employ ResNet-18 \cite{he2016deep}, and for the CIFAR100 dataset, we use ResNet-20 \cite{he2016deep}, training these models from scratch. All models are optimized using stochastic gradient descent (SGD) with a momentum of 0.9. The learning rate is set to 0.1 for the \textit{mini}ImageNet and CIFAR100 datasets and 0.002 for the CUB200 dataset. We use a mini-batch size of 64 for 120 epochs when training on the CUB200 dataset and 600 epochs for the CIFAR100 and \textit{mini}ImageNet datasets. In incremental sessions, we limit the training to 10 epochs to mitigate overfitting. We employ a data augmentation strategy where $\mathcal{M} = 1$, creating one augmented version for each image. This augmentation involves a 180° rotation combined with RGB color permutation. Set the $\delta$ coefficient as 0.5 across all datasets to ensure the new feature vector is a balanced mixture of two different augmented image feature vectors.

\subsection{Comparison with State-of-The-Arts}

 We compare our FACL framework with three FSCIL benchmark methods, CIL methods such as iCaRL\cite{rebuffi2017icarl}, EEIL \cite{castro2018end}, and Rebalancing \cite{hou2019learning}, and FSCIL-specific approaches like TOPIC \cite{tao2020few}. Additionally, we evaluate our framework against frozen-backbone FSCIL methods including SPPR \cite{zhu2021self}, F2M \cite{shi2021overcoming}, CEC \cite{zhang2021few}, FACT \cite{zhou2022forward}, TEEN \cite{wang2024few}, and SAVC \cite{song2023learning}. We include results from a basic 'finetune' \cite{song2023learning} strategy for comparative purposes. We present the performance results of our evaluation over the benchmark datasets: CUB200, \textit{mini}ImageNet, and CIFAR100. Our FACL framework effectively separates the class boundaries of the base session and adapts to future classes. On the CIFAR100 dataset, our method performed best, with a \textbf{4.77\%} average accuracy improvement and a \textbf{2.87\%} final session improvement, detailed results are provided in Table~\ref{tab4}. For the \textit{mini}ImageNet dataset, we achieve a \textbf{0.6\%} average accuracy improvement as shown in Table~\ref{tab3}. On the CUB200 dataset, we observe an average accuracy improvement of \textbf{1.59\%}, with a final session improvement of \textbf{2.2\%}, as detailed in Table \ref{tab2}. We achieve higher accuracy in the base session compared to other comparable methods. However, it is expected for our method to show a higher performance drop rate, as we focus on the base session to create feature space for future class adaptation. As a result, our method retains previous class information more effectively. Therefore, the performance drop metric is not an appropriate measure for evaluating the effectiveness of our approach. Our method outperforms the current state-of-the-art methods across all datasets.
\begin{table}[h]
\centering
\resizebox{0.48\textwidth}{!}{ % Adjust the width as needed
\begin{tabular}{@{\hskip 0.08in}l@{\hskip 0.08in}c@{\hskip 0.08in}c@{\hskip 0.08in}c@{\hskip 0.08in}c@{\hskip 0.08in}c@{\hskip 0.08in}c@{\hskip 0.08in}c@{\hskip 0.08in}c@{\hskip 0.08in}c@{\hskip 0.08in}c@{\hskip 0.08in}c@{\hskip 0.08in}c@{\hskip 0.08in}}
\toprule
\textbf{Method} & \multicolumn{9}{c}{\textbf{Accuracy in each session (\%) $\uparrow$}} & \makecell{\textbf{Average} \\ \textbf{Acc.}} & $\mathbf{\Delta{\text{FI}}}$ & \textbf{PD} \\ \cmidrule(lr){2-10}
 & 0 & 1 & 2 & 3 & 4 & 5 & 6 & 7 & 8 &  &  &  \\ \midrule
Finetune \cite{tao2020few} & 64.10 & 39.61 & 15.37 & 9.80 & 6.67 & 3.80 & 3.70 & 3.14 & 2.65 & 16.54 & -- & 61.45 \\
iCaRL \cite{rebuffi2017icarl} & 64.10 & 53.28 & 41.69 & 34.13 & 27.93 & 25.06 & 20.41 & 15.48 & 13.73 & 32.87 & +11.08 & 50.37 \\
EEIL \cite{castro2018end} & 64.10 & 53.11 & 43.71 & 35.15 & 28.96 & 24.98 & 21.01 & 17.26 & 15.85 & 33.79 & +13.20 & 48.25 \\
Rebalancing \cite{hou2019learning} & 64.10 & 53.05 & 43.96 & 36.97 & 31.61 & 26.73 & 21.23 & 16.78 & 13.54 & 34.22 & +10.89 & 50.56 \\
TOPIC \cite{tao2020few} & 64.10 & 55.88 & 47.07 & 45.16 & 40.11 & 36.38 & 33.96 & 31.55 & 29.37 & 42.62 & +26.72 & 34.73 \\
SPPR \cite{zhu2021self} & 63.97 & 65.86 & 61.31 & 57.60 & 53.39 & 50.93 & 48.27 & 45.36 & 43.32 & 54.45 & +40.67 & 20.65 \\
F2M \cite{shi2021overcoming} & 64.71 & 62.05 & 59.01 & 55.58 & 52.55 & 49.92 & 48.08 & 46.28 & 44.67 & 53.65 & +42.02 & {20.04} \\
CEC \cite{zhang2021few} & 73.07 & 68.88 & 65.26 & 61.19 & 58.09 & 55.57 & 53.22 & 51.34 & 49.14 & 59.53 & +46.49 & 23.93 \\
MetaFSCIL \cite{chi2022metafscil} & 74.50 & 70.10 & 66.84 & 62.77 & 59.48 & 56.52 & 54.36 & 52.56 & 49.97 & 60.79 & +47.32 & 24.53 \\
FACT \cite{zhou2022forward} & 74.60 & 72.09 & 67.56 & 63.52 & 61.38 & 58.36 & 56.28 & 54.24 & 52.10 & 62.24 & +49.45 & 22.50 \\
TEEN \cite{wang2024few} & 74.92 & 72.65 & 68.74 & 65.01 & 62.01 & 59.29 & 57.90 & 54.76 & 52.64 & 63.21 & +49.99 & 22.28 \\
LIMIT \cite{zhou2022few} & 73.81 & 72.09 & 67.87 & 63.89 & 60.70 & 57.77 & 55.67 & 53.52 & 51.23 & 61.84 & +48.58 & 22.58 \\
ILAR~\cite{yoon2023incrementally} & 77.50 & 73.20 & 70.80 & 66.70 & 64.00 & 62.10 & 60.50 & 58.70 & 56.40 & 65.54 & +53.75 & 21.10 \\
ALICE \cite{peng2022few}& 79.00 & 70.50 & 67.10 & 63.40 & 61.20 & 59.20 & 58.10 & 56.30 & 54.10 & 63.21 & +51.45 & 24.90 \\
MICS~\cite{kim2024mics} & 78.18 & 73.49 & 68.97 & 65.01 & 62.25 & 59.34 & 57.31 & 55.11 & 52.94 & 63.62 & +50.29 & 25.24 \\
SAVC \cite{song2023learning} & 78.77 & 73.31 & 69.31 & 64.93 & 61.70 & 59.25 & 57.13 & 55.19 & 53.12 & 63.63 & +50.47 & 25.65 \\ 
SAGG \cite{chen2024sharpness} & 79.13 & 74.68 & 71.29 & 66.98 & 64.39 & 61.35 & 59.57 & 57.93 & 55.33 & 65.63 & +52.68 & 23.80 \\ \midrule
\textbf{FACL (Ours)} & \textbf{86.20} & \textbf{81.55} & \textbf{76.95} & \textbf{72.50} & \textbf{68.75} & \textbf{65.68} & \textbf{63.16} & \textbf{60.62} & \textbf{58.20} & \textbf{70.40} & \textbf{+55.55} & {28.00} \\ \bottomrule
\end{tabular}
}
\caption{Performance Comparison of State-of-the-Art Methods on the CIFAR100 Dataset. \textbf{Average Acc.}: The average accuracy across all sessions. $\mathbf{\Delta{\text{FI}}}$: The final improvement in performance compared to the fine-tune baseline in the last session. \textbf{PD}: The performance drop is defined as the difference between the first and last session.}
\vspace{-0.9em}
\label{tab4}
\end{table}

\begin{table*}[h]
\centering
\resizebox{0.80\textwidth}{!}{ % Adjust the width as needed
\begin{tabular}{@{}lcccccccccccccc@{}}
\toprule
\textbf{Method} & \multicolumn{11}{c}{\textbf{Accuracy in each session (\%) $\uparrow$}} & \makecell{\textbf{Average} \\ \textbf{Acc.}} & $\mathbf{\Delta{\text{FI}}}$ & \textbf{PD} \\ \cmidrule(lr){2-12}
 & 0 & 1 & 2 & 3 & 4 & 5 & 6 & 7 & 8 & 9 & 10 &  &  &  \\ \midrule
Finetune ~\cite{tao2020few} & 68.68 & 43.70 & 25.05 & 17.72 & 18.08 & 16.95 & 15.10 & 10.06 & 8.93 & 8.93 & 8.47 & 21.97 & -- & 60.21 \\
iCaRL \cite{rebuffi2017icarl} & 68.68 & 52.60 & 48.61 & 44.16 & 36.62 & 29.52 & 27.83 & 26.26 & 24.01 & 23.89 & 21.16 & 36.27 & +12.69 & 47.52 \\
EEIL \cite{castro2018end} & 68.68 & 53.63 & 47.91 & 44.20 & 36.30 & 27.46 & 25.93 & 24.70 & 23.95 & 24.13 & 22.11 & 36.27 & +13.64 & 46.57 \\
TOPIC \cite{tao2020few} & 68.68 & 62.49 & 54.81 & 49.99 & 45.25 & 41.40 & 38.35 & 35.36 & 32.22 & 28.31 & 26.28 & 43.92 & +17.81 & 42.40 \\
Rebalancing \cite{hou2019learning} & 68.68 & 57.12 & 44.21 & 28.78 & 26.15 & 25.66 & 24.62 & 21.52 & 20.12 & 20.06 & 19.87 & 32.44 & +11.40 & 48.81 \\
SPPR \cite{zhu2021self} & 68.68 & 61.85 & 57.43 & 52.68 & 50.19 & 46.88 & 44.65 & 43.07 & 40.17 & 39.63 & 37.33 & 49.32 & +28.86 & 31.35 \\
F2M \cite{shi2021overcoming} & 81.07 & 78.16 & 75.57 & 72.89 & 70.16 & 68.17 & 67.01 & 65.26 & 63.36 & 61.76 & 60.26 & 69.42 & +51.79 & 20.81 \\
CEC \cite{zhang2021few} & 75.85 & 71.94 & 68.50 & 63.50 & 62.43 & 58.27 & 57.73 & 55.81 & 54.83 & 53.52 & 52.28 & 61.33 & +43.81 & 23.57 \\
FACT \cite{zhou2022forward} & 75.90 & 73.23 & 70.84 & 66.13 & 65.56 & 62.15 & 61.74 & 59.83 & 58.41 & 57.89 & 56.94 & 64.42 & +48.47 & 18.96 \\ 
TEEN \cite{wang2024few} & 77.26 & 76.13 & 72.81 & 68.16 & 67.77 & 64.40 & 63.25 & 62.29 & 61.19 & 60.32 & 59.31 & 66.63 & +50.84 & 17.95 \\ 

ALICE~\cite{peng2022few} & 77.40 & 72.70 & 70.60 & 67.20 & 65.90 & 63.40 & 62.90 & 61.90 & 60.50 & 60.60 & 60.10 & 65.75 & +51.63 & 17.30 \\ 
MICS~\cite{kim2024mics} & 78.77 & 75.37 & 72.30 & 68.72 & 67.45 & 65.40 & 64.72 & 63.39 & 61.89 & 61.89 & 61.37 & 67.39 & +52.90 & 17.40 \\
SAGG \cite{chen2024sharpness} & 81.19 & 78.12 & 75.11 & 70.62 & 70.48 & 67.18 & 66.27 & 64.72 & 63.91 & 63.42 & 62.46 & 69.41 & +53.99 & 18.73 \\ 
SAVC \cite{song2023learning} & {81.85} & {77.92} & {74.95} & {70.21} &{69.96} & {67.02} & {66.16} & {65.30} & {63.84} & {63.15} & {62.50} & 69.35 & {+54.03} & 19.35 \\ \midrule
\textbf{FACL (Ours)} & \textbf{82.74} & \textbf{80.09} & \textbf{76.89} & \textbf{71.31} & \textbf{70.51} & \textbf{68.12} & \textbf{67.54} & \textbf{66.97} & \textbf{66.05} & \textbf{65.39} & \textbf{64.70} & \textbf{70.94} & \textbf{+56.23} & {18.04} \\ \bottomrule
\end{tabular}
}
\caption{Performance Comparison of State-of-the-Art Methods on the CUB200 Dataset. \textbf{Average Acc.}: The average accuracy across all sessions. $\mathbf{\Delta{\text{FI}}}$: The final improvement in performance compared to the fine-tune baseline in the last session. \textbf{PD}: The performance drop is defined as the difference between the first and last session.}
\label{tab2}
% \vspace{-2em}
\end{table*}

\subsection{Visualization}
\begin{figure}[h]
\vspace{-1em}
  \centering
  \begin{subfigure}[t]{0.48\linewidth}
    \centering
    \includegraphics[width=\textwidth]{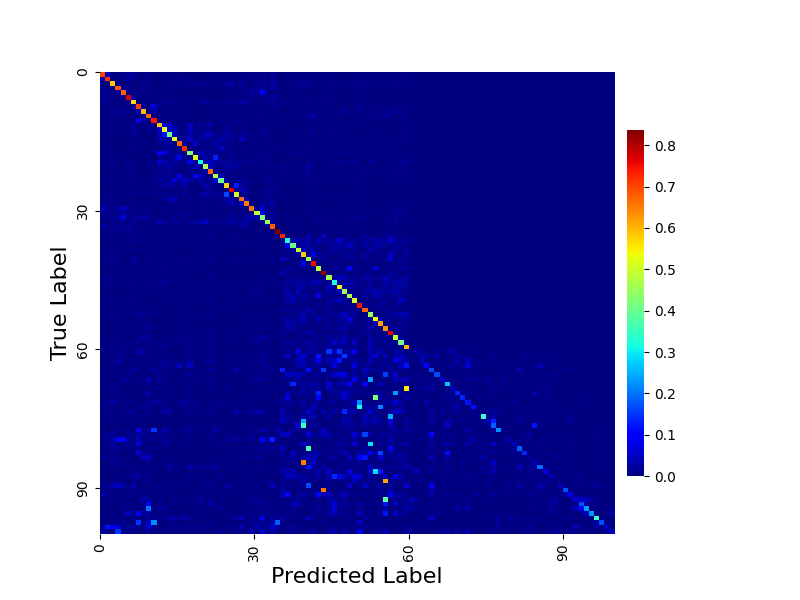}
    \caption{Baseline}
    \label{fig:short-a}
  \end{subfigure}
  \hspace{-0.05\linewidth} 
  % \hfill
  \begin{subfigure}[t]{0.48\linewidth}
    \centering
    \includegraphics[width=\textwidth]{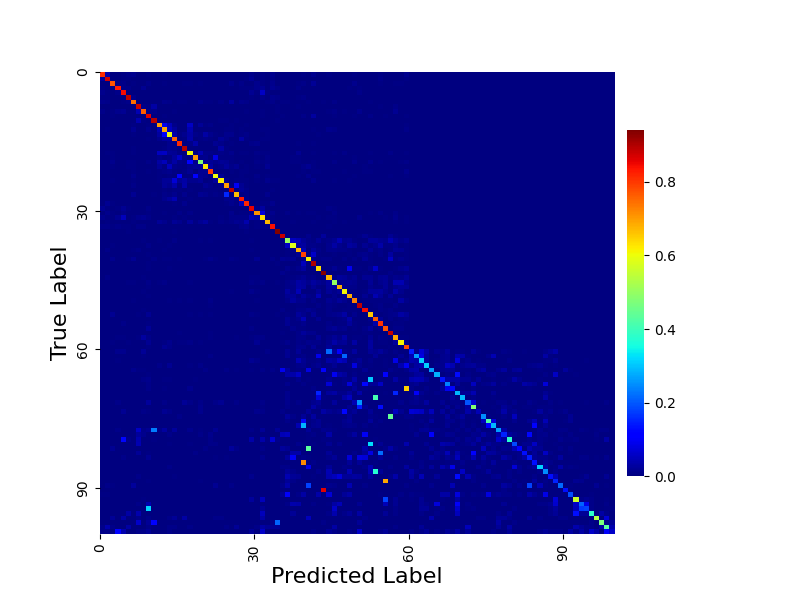}
    \caption{FACL}
    \label{fig:short-b}
  \end{subfigure}
  \vspace{-0.5em}
  \caption{Comparison of CE and FACL Confusion Matrices on the \textit{mini}ImageNet.}
  \label{fig:confusion}
  \vspace{-1em}
\end{figure}

\vspace{-0.5em}
\noindent\textbf{{Confusion Matrix:}} The confusion matrix for the baseline CE and FACL methods on the \textit{mini}ImageNet datasets is plotted, as shown in Figure \ref{fig:confusion}. The bright diagonal in the confusion matrix indicates high accuracy. The confusion matrix for the CE method demonstrates that it does not perform well on incremental classes. Conversely, the confusion matrix for our FACL method shows better performance for both the base and incremental sessions, indicating improved accuracy and fewer classification errors.
% -----------------------------------------------------------------------------------------------------------------------------------------------------------------------------
% ------------------------------------------------------------------------------------
\vspace{1em}

\begin{figure}[h]
  \centering
  \begin{subfigure}[t]{0.48\linewidth}
    \centering
    \includegraphics[width=\textwidth]{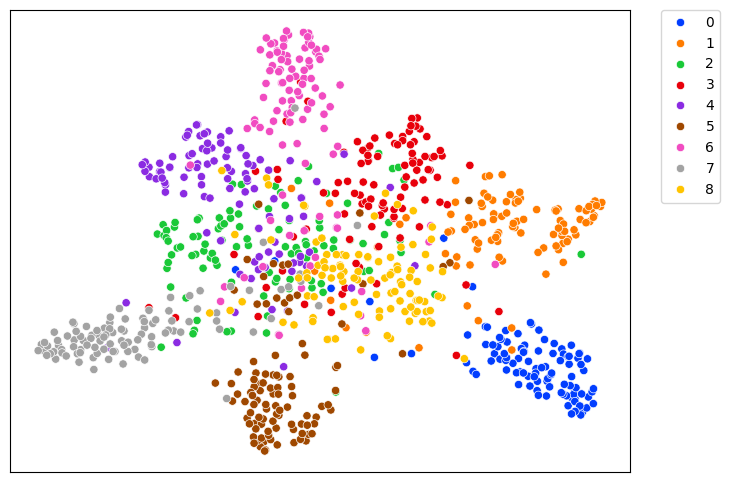}
    \caption{CE}
    \label{fig1}
  \end{subfigure}
  \hspace{0.02\linewidth}
  \begin{subfigure}[t]{0.48\linewidth}
    \centering
    \includegraphics[width=\textwidth]{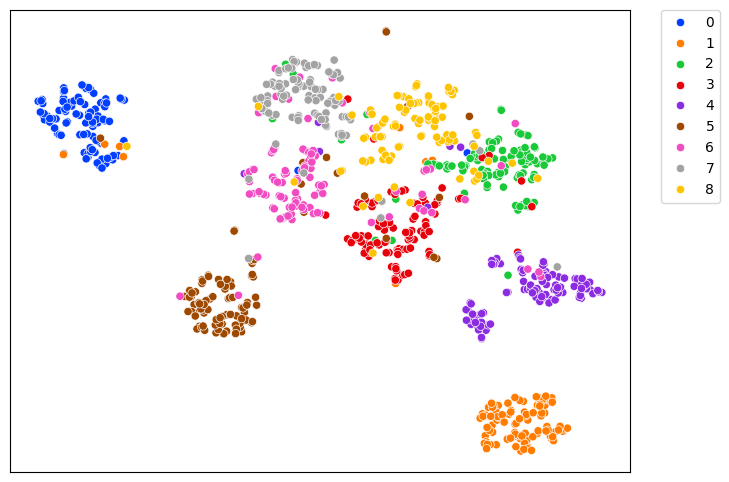}
    \caption{FACL}
    \label{fig2}
  \end{subfigure}
    \caption{Comparison of t-SNE plots between the existing framework and our proposed FACL model on the \textit{mini}ImageNet dataset (best view in color).}
  \label{fig:tsne}
  \vspace{-1em}
\end{figure}

\noindent\textbf{{Visualization of Feature Space:}}
We utilize t-SNE \cite{van2008visualizing} to visualize the feature space. In our study, six classes from the base session (0-5) and three classes from the incremental session (6-8) are randomly selected. Figure \ref{fig:tsne} demonstrates that when using CE, the classes overlap. In contrast, our feature augmentation method creates distinct spaces in the base session, allowing for the seamless incorporation of new classes.

% ----------------------------------------------------------------------------------------------------------

\begin{table}[h]
\centering
\resizebox{0.45\textwidth}{!}{ % Adjust the width as needed
\begin{tabular}{@{}lcccccccccccc@{}}
\toprule
\textbf{Method} & \multicolumn{9}{c}{\textbf{Accuracy in each session (\%) $\uparrow$}} & \makecell{\textbf{Average} \\ \textbf{Acc.}} & $\mathbf{\Delta{\text{FI}}}$ & \textbf{PD} \\ \cmidrule(lr){2-10}
 & 0 & 1 & 2 & 3 & 4 & 5 & 6 & 7 & 8 & \\ \midrule
Finetune \cite{tao2020few} & 61.31 & 27.22 & 16.37 & 6.08 & 2.54 & 1.56 & 1.93 & 2.60 & 1.40 & 13.45 & -- & 59.91 \\
iCaRL \cite{rebuffi2017icarl} & 61.31 & 46.32 & 42.94 & 37.63 & 30.49 & 24.00 & 20.89 & 18.80 & 17.21 & 33.29 & +15.81 & 44.10 \\
EEIL \cite{castro2018end} & 61.31 & 46.58 & 44.00 & 37.29 & 33.14 & 27.12 & 24.10 & 21.57 & 19.58 & 34.97 & +18.18 & 41.73 \\
Rebalancing \cite{hou2019learning} & 61.31 & 47.80 & 39.31 & 31.91 & 25.68 & 21.35 & 18.67 & 17.24 & 14.17 & 30.83 & +12.77 & 47.14 \\
TOPIC \cite{tao2020few} & 61.31 & 50.09 & 45.17 & 41.16 & 37.48 & 35.52 & 32.19 & 29.46 & 24.42 & 39.64 & +23.02 & 36.89 \\
IDLvQ-C \cite{chen2020incremental} & 64.77 & 59.87 & 55.93 & 52.62 & 49.88 & 47.55 & 44.83 & 43.14 & 41.84 & 51.16 & +40.44 & 22.93 \\
SPPR \cite{zhu2021self} & 61.45 & 63.80 & 59.53 & 55.53 & 52.50 & 49.60 & 46.69 & 43.79 & 41.92 & 52.76 & +40.52 & 19.53 \\
F2M \cite{shi2021overcoming} & 67.28 & 63.80 & 60.38 & 57.06 & 54.08 & 51.39 & 48.82 & 46.58 & 44.65 & 54.89 & +43.25 & 22.63 \\
CEC \cite{zhang2021few} & 72.00 & 66.83 & 62.97 & 59.43 & 56.70 & 53.73 & 51.19 & 49.24 & 47.63 & 57.75 & +46.23 & 24.37 \\
MetaFSCIL \cite{chi2022metafscil} & 72.04 & 67.94 & 63.77 & 60.29 & 57.58 & 55.16 & 52.90 & 50.79 & 49.19 & 58.85 & +47.79 & 22.85 \\
FACT \cite{zhou2022forward} & 72.56 & 69.63 & 66.38 & 62.77 & 60.60 & 57.33 & 54.34 & 52.16 & 50.49 & 60.70 & +49.09 & 22.07 \\
LIMIT \cite{zhou2022few} & 72.32 & 68.47 & 64.30 & 60.78 & 57.95 & 55.07 & 52.70 & 50.72 & 49.19 & 59.06 & +47.79 & 23.13 \\
TEEN \cite{wang2024few} & 73.53 & 70.55 & 66.37 & 63.23 & 60.53 & 57.95 & 55.24 & 53.44 & 52.08 & 61.43 & +50.68 & 21.45 \\
SAGG \cite{chen2024sharpness} & 78.77 & 74.75 & 71.20 & 68.12 & 65.16 & 62.10 & 59.22 & 57.28 & 56.19 & 65.89 & +54.79 & 22.58 \\ 
ALICE \cite{peng2022few} & 80.60 & 70.60 & 67.40 & 64.50 & 62.50 & 60.00 & 57.80 & 56.80 & 55.70 & 63.99 & +54.30 & 24.90 \\
SAVC \cite{song2023learning} & 81.12 & 76.14 & 72.43 & 68.92 & 66.48 & 62.95 & 59.92 & 58.39 & 57.11 & 67.05 & +55.71 & 24.01 \\
MICS~\cite{peng2022few} & 84.40 & 79.48 & 75.09 & 71.40 & 68.89 & \textbf{66.16} & \textbf{63.57} & \textbf{61.79} & \textbf{60.74} & 70.17 & \textbf{+59.34} & 23.66 \\
\midrule
\textbf{FACL (Ours)} & \textbf{86.68} & \textbf{81.49} & \textbf{76.65} & \textbf{72.65} & \textbf{69.71} & {66.02} & {63.08} & {61.17} & {59.48} & \textbf{70.77} & {+58.08} & 27.20 \\ \bottomrule
\end{tabular}
}
\caption{Performance Comparison of State-of-the-Art Methods on the \textit{mini}ImageNet Dataset. \textbf{Average Acc.}: The average accuracy across all sessions. $\mathbf{\Delta{\text{FI}}}$: The final improvement in performance compared to the fine-tune baseline in the last session. \textbf{PD}: The performance drop is defined as the difference between the first and last session.}
\label{tab3}
\vspace{-1.5em}
\end{table}

% \vspace{-1em}
\subsection{Ablation study}
\noindent\textbf{{Impact of Framework Components:}}
We conduct several ablation studies, and use classification loss as the baseline to examine the influence of different components of our framework. The components include Cross-Entropy (CE) loss, Self-Supervised Contrastive Learning (SSCL), Proxy Classes (PC), and Feature Augmentations (FA). Initially, we perform experiments using only CE. We observe that applying CE alone in the base session did not generalize well to novel classes. Conversely, using CE in combination with SSCL and PC achieves a performance improvement of 5.74\% compared to the CE baseline. Furthermore, incorporating FA resulted in a significant enhancement during the base session, generalizing well to novel classes and achieving a 12.33\% accuracy improvement. Table \ref{tab5} presents the detailed results on the CIFAR100 dataset.
% \vspace{-1em}

\begin{figure}[h!]
    \centering
    \includegraphics[width=0.29\textwidth]{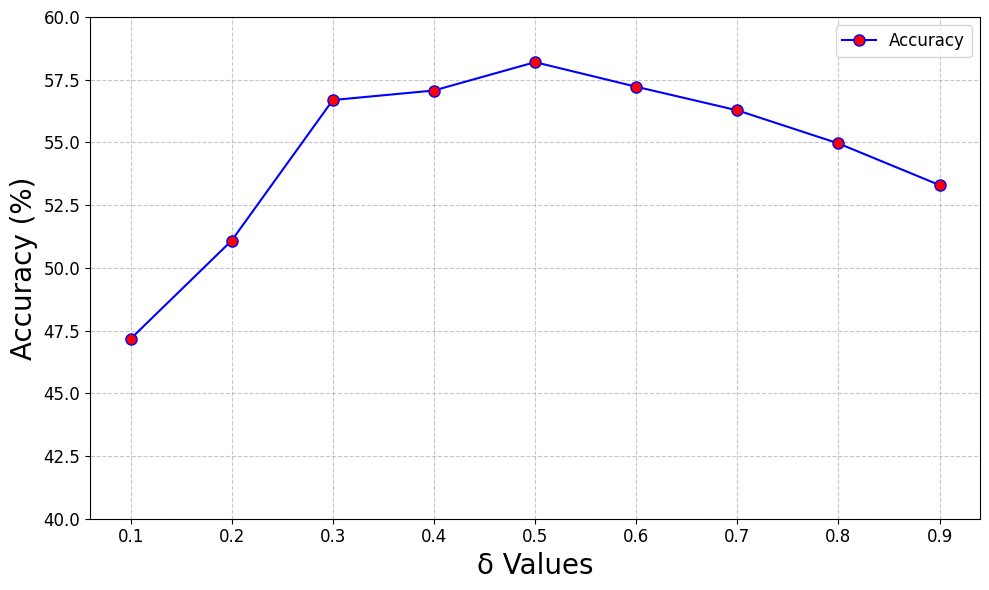} 
    \vspace{-1em}
    \caption{Accuracy values obtained during the final session for various (\(\delta\)) values on the CIFAR100 dataset.}
    \label{fig:accuracy-delta}
    \vspace{-0.5em}
\end{figure}

\begin{table}[h]
\centering
\setlength{\tabcolsep}{3pt}
\scalebox{0.6}{
\begin{tabular}{cccccccccccccc}
\toprule
\textbf{CE} & \textbf{SSCL} & \textbf{PC} & \textbf{FA} & \multicolumn{9}{c}{\textbf{Acc. in each session (\%) $\uparrow$}} & $\mathbf{\Delta{\text{FI}}}$ \\ \cmidrule(lr){5-13}
 & & & & 0 & 1 & 2 & 3 & 4 & 5 & 6 & 7 & 8 & \\ \midrule
\checkmark & & & & 74.77 & 68.34 & 63.59 & 59.40 & 55.83 & 52.95 & 50.61 & 48.14 & 45.87 & - \\
\checkmark & & \checkmark & & 74.93 & 69.66 & 65.47 & 62.28 & 59.14 & 55.95 & 53.16 & 51.13 & 49.58 & +3.71 \\
\checkmark & \checkmark & & & 78.37 & 72.92 & 67.94 & 63.91 & 60.70 & 57.68 & 55.28 & 52.65 & 50.51 & +4.64 \\
\checkmark & \checkmark & \checkmark  & & 79.08 & 73.14 & 68.39 & 64.15 & 60.78 & 57.66 & 55.60 & 53.60 & 51.61 & +5.74 \\
\checkmark &  & \checkmark  &\checkmark & 86.53 & 81.32 & 75.68 & 70.90 & 67.63 & 64.42 & 61.72 & 59.31 & 57.21 & +11.34 \\
\checkmark & \checkmark & \checkmark & \checkmark & 86.20 & 81.55 & 76.96 & 72.51 & 68.75 & 65.68 & 63.17 & 60.62 & 58.20 & +12.33 \\
\bottomrule
\end{tabular}
}

\caption{Ablation studies on the CIFAR100 benchmark, where \textbf{CE} represents Cross Entropy loss, \textbf{SSCL} stands for Self-Supervised Contrastive loss, \textbf{PC} denotes Proxy Classes, and \textbf{FA} signifies Feature Augmentation. The metric $\mathbf{\Delta{\text{FI}}}$ indicates the relative improvements of the final session compared to the CE baseline.}
\label{tab5}
\vspace{-0.5em}
\end{table}
\noindent\textbf{ {Analysis of $\delta$ Coefficient:}}
We investigate the impact of varying the delta ($\delta$) value utilized in generating a mixture of augmented image features. Reporting the accuracy obtained in the final session on the CIFAR100 dataset for different delta values. as illustrated in Figure \ref{fig:accuracy-delta}. Notably, the highest accuracy is achieved with $\delta = 0.5$ across all datasets. This particular combination ensures that the new feature is a balanced mixture of two distinct augmented image features.

\begin{table}[h]
\centering
\setlength{\tabcolsep}{3pt} % Adjust the column separation here
\resizebox{0.45\textwidth}{!}{ % Adjust the width as needed
\begin{tabular}{@{}cccccccccccc@{}}
\toprule
\textbf{Ablation Variations} & \multicolumn{9}{c}{\textbf{Accuracy in each session (\%) $\uparrow$}} & $\mathbf{\Delta{\text{FI}}}$ \\ \cmidrule(lr){2-10}
 & 0 & 1 & 2 & 3 & 4 & 5 & 6 & 7 & 8 \\ \midrule
CE & 74.77 & 68.34 & 63.59 & 59.40 & 55.83 & 52.95 & 50.61 & 48.14 & 45.87 & - \\ \midrule
$Z^i_{aug} + N(0, I)$ & 78.37 & 73.06 & 68.63 & 64.41 & 61.26 & 58.22 & 56.00 & 53.72 & 51.56 & +5.69 \\

$Z^i_{ori} + N(0, I)$ & 78.51 & 73.12 & 68.37 & 64.57 & 61.03 & 58.01 & 55.75 & 53.64 & 51.69 & +5.82\\
$Z^i_{ori} + Z^j_{aug}$ & 79.88 & 73.94 & 69.40 & 65.12 & 62.01 & 58.84 & 56.64 & 53.92 & 51.58 & +6.01\\
$Z^i_{ori} + Z^j_{ori}$ & 84.66 & 79.76 & 74.27 & 69.62 & 65.73 & 62.69 & 59.60 & 57.10 & 54.58 & +8.71\\
\textbf{$\boldsymbol{Z^i_{\text{aug}} + Z^j_{\text{aug}}}$}
 & \textbf{86.20} & \textbf{81.55} & \textbf{76.96} & \textbf{72.51} & \textbf{68.75} & \textbf{65.68} & \textbf{63.17} & \textbf{60.62} & \textbf{58.20} & \textbf{+12.33} \\
\bottomrule
\end{tabular}
}
% \vspace{0.01cm}
\caption{Ablation study results on CIFAR100 with variations of our framework, comparing the accuracy (\%) across different sessions for Gaussian Noise and Feature Mixture configurations.}
\label{tab6}
\vspace{-1em} 
\end{table}

% \vspace{-1em}
\noindent\textbf{{Impact of Random Noise:}}
Experiments are conducted to explore the impact of adding random noise for feature augmentations. Gaussian noise  \( N(0, I) \) is introduced in place of \( Z^j_{\text{aug}} \) in equation \ref{eq:augmented_features}. It is observed that the mixture of augmented image features achieves improved performance compared to using Gaussian noise alone. The comparable results are presented in Table \ref{tab6}.\\
% \vspace{-1cm}
\noindent\textbf{{Impact of Feature Mixture Variations:}}
In our method, we use the augmented version of the original image to perform feature augmentation. As part of our ablation study, we conduct experiments on various mixture variations for feature augmentations. Specifically, mixing original and augmented features ($Z^i_{ori} + Z^j_{aug}$) results in a 6.01\% improvement, whereas using original features from both images ($Z^i_{ori} + Z^j_{ori}$) yields a 8.71\% gain. However, the results in Table \ref{tab6} clearly demonstrate that using the augmented image features ${Z^i_{aug} + Z^j_{aug}}$ to perform the feature augmentations works better than using the other combinations. 
% Thus, the augmented features are more beneficial for creating the space in the base session.

\noindent\textbf{{Analysis of Data Augmentation Methods:}}
We compare our results with existing data augmentation methods, including Mixup\cite{zhang2017mixup}, CutMix\cite{yun2019cutmix}, and Manifold Mixup\cite{verma2019manifold}. Our findings demonstrate significantly improved outcomes compared to these approaches, with the results presented on the \textit{mini}ImageNet dataset in Table \ref{tab7}.

\noindent{\textbf{Discussion:}}
Existing approaches, such as ALICE \cite{peng2022few} and Ensemble\cite{zhu2024enhanced}, utilize Mixup\cite{zhang2017mixup} and CutMix\cite{yun2019cutmix} for data augmentation, where two images from different classes are combined to create a new image. However, these methods often fail to capture the semantic meaning, which is crucial for generalizing to future classes. In contrast, our method FACL, employs feature augmentation based on the augmented image, preserving "fine-grained" information. Moreover, while Moreover, other methods turn the problem into a $(\mathcal{Y} + \mathcal{Y} \times (\mathcal{Y} - 1)/2)$ class classification task, our approach maintains the same number of proxy classes as the original ones, reducing the risk of overfitting. Additionally, DBONet~\cite{guo2023decision} and MICS\cite{peng2022few} uses Manifold-MixUp~\cite{verma2019manifold} for augmentation, operating on hidden representations, which can result in a loss of semantic information and important feature values. In contrast, our method generates virtual classes directly from the features, retaining critical information.

\begin{table}[h]
\centering
\setlength{\tabcolsep}{3pt} % Adjust the column separation here
\resizebox{0.45\textwidth}{!}{ % Adjust the width as needed
\begin{tabular}{@{}cccccccccccc@{}}
\toprule
\textbf{Augmentation Methods} & \multicolumn{9}{c}{\textbf{Accuracy in each session (\%) $\uparrow$}} & $\mathbf{\Delta{\text{FI}}}$ \\ \cmidrule(lr){2-10}
 & 0 & 1 & 2 & 3 & 4 & 5 & 6 & 7 & 8 \\ \midrule
Manifold-MixUp~\cite{verma2019manifold} & 69.60 & 64.55 & 60.77 & 57.47 & 54.69 & 51.86 & 49.29 & 47.40 & 46.01 & - \\
CutMix~\cite{yun2019cutmix} & 74.90 & 69.81 & 65.64 & 62.20 & 58.93 & 55.77 & 53.33 & 51.24 & 49.20 & +3.19\\
Mixup~\cite{zhang2017mixup} & 82.30 & 76.16 & 71.24 & 68.11 & 64.74 & 61.49 & 58.38 & 55.74 & 54.83 & +8.82\\
\textbf{FACL (Ours)} 
 & \textbf{86.20} & \textbf{81.55} & \textbf{76.96} & \textbf{72.51} & \textbf{68.75} & \textbf{65.68} & \textbf{63.17} & \textbf{60.62} & \textbf{58.20} & \textbf{+12.19} \\
\bottomrule
\end{tabular}
}
% \vspace{0.01cm}
\caption{Comparative Performance of Data Augmentation Methods (Mixup, CutMix, and Manifold Mixup) on the \textit{mini}ImageNet Dataset.}
\label{tab7}
\vspace{-2.5em} 
\end{table}

%-------------------------------------------------------------------------

\vspace{-1em}
\section{Conclusion}
In this paper, we address the FSCIL problem by introducing a Feature Augmentation via Contrastive Learning (FACL) framework. Existing frameworks often fail to effectively separate the base classes, leading to new classes collapsing into the base classes. To address this issue, we use feature augmentation to create proxy classes. These proxy classes expand the base session, allowing future classes to be easily adapted without confusion with the base classes. Employing the FACL framework establishes a robust base session for FSCIL, achieving SOTA performance on all three FSCIL benchmarks.\\
\vspace{-1em}
\section*{Acknowledgment}
Parinita acknowledges the support of the University Grants Commission (UGC), Ministry of Education, India, through the Junior Research Fellowship (JRF), which contributed significantly to the completion of this work.

\newpage
%%%%%%%%% REFERENCES
{\small
\bibliographystyle{ieee_fullname}
\bibliography{egbib}
}

\end{document}